\documentclass[runningheads]{llncs}
\usepackage{graphicx}
\usepackage{amsfonts}
\usepackage{amsmath}
\usepackage{amssymb}
\usepackage{diagbox}
\begin{document}
\title{Feature Super-Resolution Based Facial Expression Recognition for Multi-scale Low-Resolution Faces}
\titlerunning{Feature Super-Resolution Based FER for Multi-scale Low-Resolution Faces}
%
\author{Wei Jing \inst{1} \and
Feng Tian \inst{1} \and
Jizhong Zhang \inst{1} \and
Kuo-Ming Chao \inst{2} \and
Zhenxin Hong \inst{1} \and
Xu Liu \inst{1}
}
%
\authorrunning{Wei Jing et al.}
%
\institute{National Engineering Lab of Big Data Analytics, Faculty of Electronic and Information Engeneering, Xi'an Jiaotong University, China \and
School of Engennering and Computing, Coventry University, Coventry, UK\\
\email{\{a351111798, zjz19960805, hoho9713\}@stu.xjtu.edu.cn,  fengtian@mail.xjtu.edu.cn, csx240@coventry.ac.uk}}


%
\maketitle              
\begin{abstract}
Facial Expressions Recognition(FER) on low-resolution images is necessary for applications like group expression recognition in crowd scenarios(station, classroom etc.). Classifying a small size facial image into the right expression category is still a challenging task. The main cause of this problem is the loss of discriminative feature due to reduced resolution. Super-resolution method is often used to enhance low-resolution images, but the performance on FER task is limited on images of very low resolution. In this work, inspired by feature super-resolution methods for object detection, we proposed a novel generative adversary network-based feature level super-resolution method for robust facial expression recognition (FSR-FER). In particular, a pre-trained FER model was employed as a feature extractor, and a generator network G and a discriminator network D are trained with features extracted from low-resolution and original high-resolution images. Generator network G tries to transform features of low-resolution images to more discriminative ones by making them closer to the ones of corresponding high-resolution images. For better classification performance, we also proposed an effective classification-aware loss re-weighting strategy based on the classification probability calculated by a fixed FER model to make our model focus more on samples that are prone to misclassifiy. Experimental results on Real-World Affective Faces (RAF) Database demonstrate that our method achieves satisfying results on various down-sample factors with a single model and has better performance on low-resolution images compared with methods using image super-resolution and expression recognition separately.

\keywords{Facial expression recognition  \and Feature super-resolution \and Generative Adversarial Network \and Various low resolutions}
\end{abstract}
 \begin{figure}
\includegraphics[width=\textwidth]{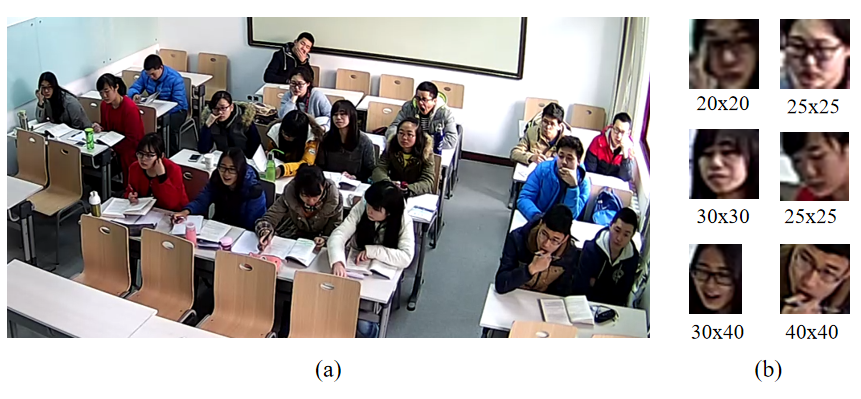}
\caption{The example of classroom review system, (a) the example image of whole classroom. (b) facial images of various resolutions marked below } \label{fig_class}
\end{figure}

\section{Introduction}
Expression recognition is of great significance for some practical applications of education and healthcare, etc. For example, by identifying student expressions from classroom records, it can help educators to better understanding of the teaching effect\cite{wei2017bnu}. Automatic pain recognition in the medical field has also received more and more attention\cite{wang2017pain,haque2018painsurvey}.

Recent expression recognition methods include CNN-based backbone networks for feature extraction, and expression classification results are obtained through fully connected layers or using SVM and other methods. Thanks to the powerful feature extraction capabilities of ResNet\cite{he2016resnet}, Inception network\cite{szegedy2015going} and other networks, it can usually achieve a satisfactory result on most data sets. In real-world scenarios, especially in crowded scenes, facial pictures come from different low resolutions. For example, in the classroom review system, as shown in Figure ~\ref{fig_class}, most of the students' facial images are less than 50x50 pixels, and the smaller images are less than 20x20 pixels, which is much smaller than the commonly used data sets. Reduced image resolution will not only make tasks such as object detection difficult\cite{noh2019better},but also result in accuracy decreasing of facial expression recognition according to our experiment results in section 4. 

Image super-resolution(ISR) technology can recover high-resolution images with rich details from low-resolution images. In recent years, a large number of image super-resolution algorithms\cite{zhang2018rdn,lim2017edsr,dai2019second,hu2019meta,zhang2018rcan,lai2017deep} have been proposed. They can help restore image details and have extraordinary performances on PSNR, SSIM and other indicators. Some studies\cite{liu2020facial,cheng2017robust} utilize image super-resolution methods to enhance low-resolution images to obtain better results.It is indicated in\cite{tan2018feature} that feature information loss due to reduced image resolution leads to performance degradation on specific tasks, and  feature-level super-resolution has achieved satisfactory results on small object detection. Inspired by feature-level super-resolution works\cite{tan2018feature,noh2019better,li2017perceptual}, we propose a feature super-resolution framework for facial expression recognition, and compared it with the methods of directly using popular image super-resolution. In which, firstly, we use a pre-trained CovPoolFER\cite{acharya2018covariance} model as a feature extractor to extract expression features from high-resolution images and downsample low-resolution images. Next, as a GAN-based model, a generator network G takes the features of  low-resolution images as input and generate highly discriminative features. Different from most GAN models, here discriminator network D does not estimate the probabilities that feature coming from high-resolution images but measures the difference between high and low-resolution image features. Through alternative training, G tries to narrow this gap so that the generated features are closer to the real ones in distribution. Besides, considering that the ultimate goal is to perform correct classification, we propose a perceptual loss and a simple but effective classification-aware loss re-weighting strategy to make the model on samples that are difficult to be correctly classified.

In summary, the main contributions of this work are:
(1)We propose a facial expression recognition method based on feature super-resolution that can be used for various low-resolution inputs. (2)We introduced a new class-aware loss re-weighting strategy for GAN training, helping our model focus on samples not only with large loss but hard to be classified into corresponding category. (3)Experiments are conducted to compare our method with methods using ISR and FER separately. Results show that our method has better performance on low-resolution images.

\section{Related Work}
Facial Expression Recognition has been investigated for a long time, numerous methods\cite{li2017rafdb,ding2017facenet2expnet,yang2018deexp,zhang2018joint}have been studied. Studies on low-resolution facial expression recognition are few. Some researches utilize image super-resolution method to restore high-resolution images or carry ISR and FER jointly. Here we focus on static images FER and super-resolution methods.
\subsection{Static Image Facial Expression Recognition}
To achieve good expression recognition performance, most of recent facial expression recognition methods employ standard CNN-based networks such as Residual networks, Inception networks and VGG networks to extract representative features\cite{hasani2017resnetexample,zhao2016peak,ding2017facenet2expnet}. Generally, these networks are often pre-trained on standard datasets related to FER or face recognition , such as FER-2013 and LFW\cite{huang2008lfw} etc. After feature extraction, supervised classifier(e.g SVM\cite{hesse2012svm}, softmax\cite{jung2015softmax}) are trained to classify extracted feature into right categories. In some recent researches \cite{huang2017riemannian,yu2017covpoolcnn,dai2019second}, second-order feature statistics show its benefits on different image tasks. In \cite{acharya2018covariance}, authors introduced covariance pooling operation to extract second-order statistics and reported a state-of-the-art performance on RAF-DB. When it come to low-resolution FER, super-resolution methods are often used to increase resolution and achieve better performance.In\cite{cheng2017robust}, authors proposed an encoder-decoder framework to do ISR and FER jointly. In\cite{liu2020facial}, authors proposed a graph convolution networks based super-resolution framework for facial expression restoration.
\subsection{Super-Resolution}
For single-image super-resolution (SISR), the goal is to recover high-resolution images from low-resolution images. SRCNN\cite{dong2015srcnn} model is a fully convolutional network model that directly learns the mapping of low-resolution images to high-resolution images through end-to-end training. Following this pioneering work, a large number of methods are proposed with improvements on network structure\cite{lim2017edsr,zhang2018rdn} and popular attention mechanism\cite{zhang2018rcan,dai2019second}. Different from most existing works that train a specific model for each scale factor which, in \cite{hu2019meta}, authors proposed a magnification-arbitrary method.

Image super-resolution helps to increase resolution and restore details in images, and has a certain degree of contribution to the performance improvement in some works\cite{liu2020facial,cheng2017robust}. However, image super-resolution itself is time consuming and it’s hard for image super-resolution to recover the details at very low resolution, which means it cannot guarantee to recovered features that are  discriminative enough for specific tasks(e.g object detection, expression classification .etc). For small objection detection, Li et al.\cite{li2017perceptual} proposed perceptual GAN to generate super-resolved features for a small instance, which utilizes high-resolution target features as supervision signals and learn the residual between high and low-resolution image features.The work of authors\cite{tan2018feature} demonstrate that FSR performs better than ISR on recovery of task-relative features and a feature-level L2 loss can help generator produce more discriminative features with faster convergence. However, method in \cite{tan2018feature} has limited improvement of classification accuracy on FER task.

\section{Proposed Methods}
Our main goal is to improve the expression recognition effect on low-resolution face images. In this section, we first introduce the network architecture of the proposed methods. And then, we discuss the loss functions of the feature generator. At last we describe the classification-aware loss re-weighting strategy. 
\subsection{Network Architecture}
 \begin{figure}
\includegraphics[width=\textwidth]{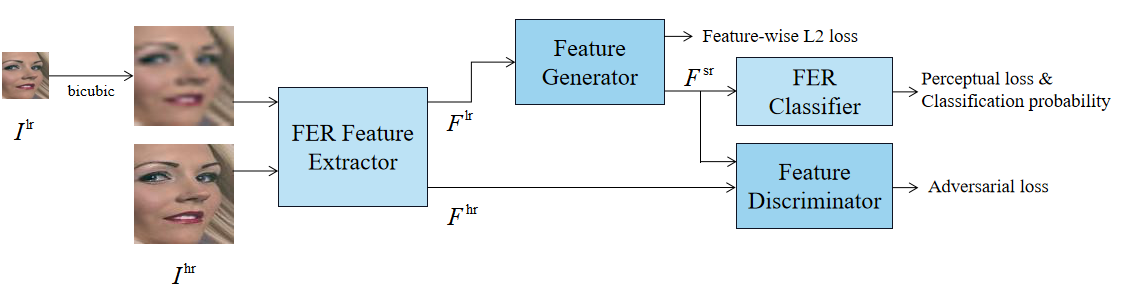}
\caption{Overall architecture of our feature super-resolution facial expression recognition (FSR-FER) approach. Fixed feature extractor and classifier are split from a FER model. Generator G tries to produce highly discriminative features by adversarial training with discriminator D. FSR classifier here is utilized to provide perceptual loss and also serves as a supervisor in proposed classification-aware loss re-weighting strategy.} \label{fig_arch}
\end{figure}
Our model consists of three main parts ,i.e., a baseline FER model which is further divided into two parts for feature extraction and perceptual loss calculation, feature generator network,and discriminator. The architecture of our model is shown in Figure ~\ref{fig_arch}.The input high-resolution image and low-resolution image are denoted by $\textbf{I}^{hr}$ and $\textbf{I}^{lr}$ respectively, and the corresponding extracted facial features are $\textbf{F}^{hr}$ and $\textbf{F}^{lr}$. In  our approach, we employ the CovPoolFER\cite{acharya2018covariance} model, $\textbf{F}^{hr}$ and $\textbf{F}^{lr}$ are the obtained second-order features,which can be written as follow:
\begin{equation}
    F^{hr}=SPD(F(I^{hr}))
\end{equation}
\begin{equation}
    F^{lr}=SPD(F(I^{lr}))
\end{equation}
where $\textit{F()}$ denotes the CNN network and $\textit{SPD()}$ refers to the operations of covariance pooling and SPDNet layers. Covariance pooling is for capturing second-order information. SPDNet layers further flatten second-order features to be able to apply standard loss functions. Operation details can be found in\cite{huang2017riemannian,yu2017covpoolcnn}.  

After feature extraction, similar to \cite{tan2018feature}, the feature generator G tries to transform $\textbf{F}^{lr}$ to a more discriminative new feature $\textbf{F}^{sr}$, which can be defined as follow:
\begin{equation}
    F^{sr}=G(F^{lr})
\end{equation}
G aims is to make generated $\textbf{F}^{sr}$ more discriminative, which means that G needs to learn the mapping from $\textbf{F}^{lr}$ to corresponding $\textbf{F}^{hr}$ from high-resolution images.According to the idea of GAN, the discriminator is trained to distinguish $\textbf{F}^{sr}$ and $\textbf{F}^{hr}$. We observe that simply training a pair of G and D has a limited improvement on the final expression classification accuracy. Only feature-wise L2 loss\cite{tan2018feature} is not sufficient for facial expression recognition, so we employ CovPoolFER model again as an extra supervisor to cast a stronger constrain on the training of G.To achieve this, we propose a perceptual loss for generator. Besides, inspired by PISA loss\cite{cao2019prime}, we propose a simple but effective classification-aware loss re-weighting strategy. This re-weighting strategy can improve the performance of proposed model by making it focus more on the easily misclassified samples.
\subsection{Loss Functions for Feature Generator}
Generative adversarial networks\cite{goodfellow2014gan}(GAN) has been widely used in the areas of image restoration super-resolution. In the idea of GAN, adversarial training is performed between a generator G and a discriminator D. Through this process, G learns to transform a given input or noise into target output distribution. This process can be formally defined as follows:
\begin{equation}
    \mathop{\min}_{G}\mathop{\max}_{D} V(D,G)=\mathbb{E}_{x\sim \mathbb{P}_{r}}[\log D(x)]+\mathbb{E}_{\tilde{x}\sim  \mathbb{P}_{g}}[1-\log D(\tilde{x})]
\end{equation}
However, training GAN following the original loss function is unstable in practical. For better training effect, we follow the idea of  WGAN-div\cite{wu2018wgandiv}. Similar to WGAN\cite{arjovsky2017wasserstein}, here discriminator D does not output the probability that the input feature belonging to the real high-resolution distribution. Instead, Wasserstein distance is utilized to measure the difference between the two diversities. The loss function of the generator and discriminator can be written as follows:
\begin{equation}
    L(G)=-\mathbb{E}_{\tilde{x} \sim \mathbb{P}_{g}}[D(\tilde{x})]
\end{equation}
\begin{equation}
    L(D)= \mathbb{E}_{\tilde{x}\sim \mathbb{P}_{g}}[D(\tilde{x})]-\mathbb{E}_{x\sim \mathbb{P}_{r}}[D(x)]
\end{equation}
In order to meet the k-Lipschitz constraint required by WGAN, methods such as gradient punishment and spectral normalization are often used in training process. However WGAN-div\cite{wu2018wgandiv} futher proposed Wasserstein divergence where the  k-Lipschitz constraint is not required. Finally the adversary loss of our G and D can be written as follows: 
\begin{equation}
    L_{gan}(G)=\mathbb{E}_{\tilde{x}\sim\mathbb{P}_{g}}[D(\tilde{x})]
\end{equation}
\begin{equation}
    L_{gan}(D)=\mathbb{E}_{x\sim \mathbb{P}_{r}}[D(x)]-\mathbb{E}_{\tilde{x}\sim \mathbb{P}_{g}}[D(\tilde{x})]+k\mathbb{E}_{\hat{x}\sim \mathbb{P}_{u}}[\|\nabla _{\hat{x}} D(\hat{x})\|^{p}]
\end{equation}
where $x$ is the real data of $F^{hr}$, and $\tilde{x}$ is generated data of $F^{sr}$. $\hat{x}$ is sampled as a linear combination of $F^{hr}$ and $F^{sr}$.

Perceptual loss\cite{johnson2016perceptual}is often used in super-resolution researches to recover high-frequency features and convergence fast to satisfactory results. In this paper, we propose a new method to calculate the perceptual loss, which is defined as follows:
\begin{equation}
    L_{p}(G)=\frac{1}{m}\sum_{i=1}^{m}\|C(F^{sr}_{i})-C(F^{hr}_{i})\|^{2}_{2}
\end{equation}
where $\textit{C()}$ refers to the part of CovPoolFER model after SPDNet layers, and its output is a 128-dimensional vector. Compared with the original perceptual loss, this loss equivalent to calculating the perceptual loss of the two features in a deeper feature space.The total loss function of generator G can be written as follow:
\begin{equation}
    L(G)=L_{gan}+L_{p}+L_{2}
\end{equation}
where $L_{2}$ refers to feature-wise L2 loss.

\subsection{Classification-Aware Loss Re-Weighting Strategy} 
Some research suggests that different samples do not contribute equally to the training process of a model. Focal loss\cite{lin2017focal} modified the cross-entropy loss to make the model focus on hard examples. In PISA\cite{cao2019prime}, classification probability is utilized to re-weight the regression loss, which makes the model pay more attention to prime samples with higher probability to classify correctly. In our task, samples that are easily misclassified contribute more in the training process. To this end, we propose a loss re-weighting strategy, which is defined as follow:

\begin{gather}
w_{i}=(\sigma-p_{i})^{r}\\
L_{re} = \sum_{i=1}^{m}w_{i}L(x_{i},x_{i}) \notag
\end{gather}
where $p_{i}$ denotes the predicted probability of the corresponding ground truth expression category. We set $\sigma$ and $r$ as positive parameters larger than 1. This re-weighting strategy, utilizing the supervised information provided by the pre-trained expression recognition model, will lead the model to pay more attention to samples that are difficult to classify correctly. In the training process, this re-weighting strategy applies on loss functions of generator.

\subsection{Implementation Details}
For GAN training, we employ the residual-in-residual dense block(RRDB) of ESRGAN\cite{wang2018esrgan} as basic blocks. We modified RRDB by reducing its internal channels, and build our G with 6 blocks. We follow the discriminator network architecture of ESRGAN but batch-normalization operation is removed for it may result in gradient explosion in our experiments. The parameter k and p are set to 2 and 6 respectively. $\sigma$ and $r$ are set to 1.5 and 1. Our model is trained by Adam optimizer \cite{kingma2014adam}, and $\beta_{1}$ is set to 0 for the stability of GAN training. The learning rate is set to $2\times10^{-4} $ and halved at [20k,50k,100k,200k] iterations. Our models are implemented with TensorFlow framework and trained on one NVIDIA Tesla V100 GPU.

\section{Experiments}
\subsection{Dataset}
\begin{figure}
 \centering
\includegraphics[width=8cm]{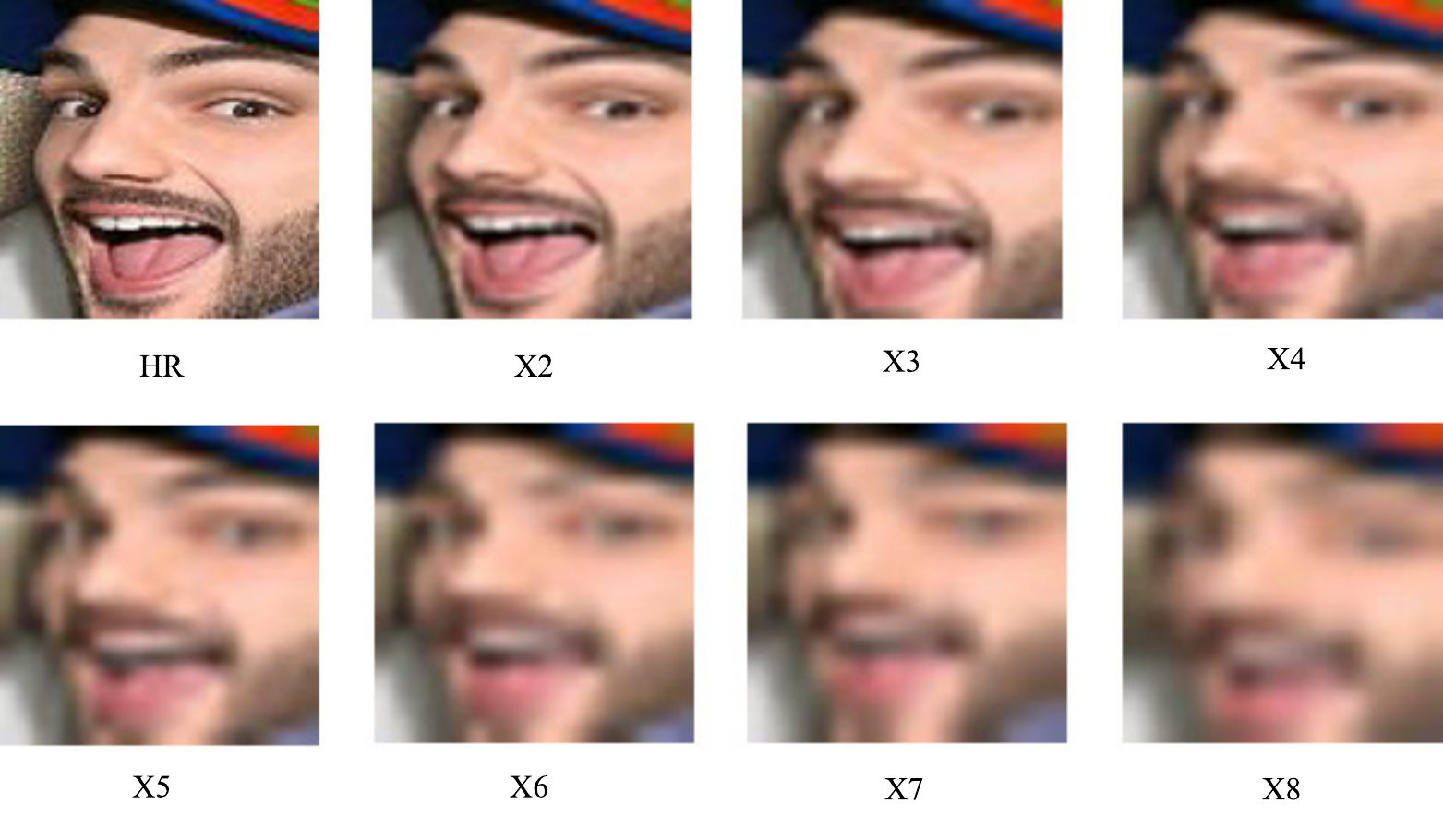}
\caption{A HR image(resolution: 100x100) from RAF-DB and down-sample results with different down-sampling factors from x2 to x8.} \label{fig_downsample}
\end{figure}

 To evaluate the expression recognition performance  on low-resolution images, RAF-DB\cite{li2017rafdb} is used in our experiment. RAF-DB was gathered with various search engines, and each image has been independently labelled by about 40 annotators. It contains 15339 images labelled with seven basic emotion categories of which 12271 are for training and the other 3068 are for validation. Like most super-resolution studies, we downsample high-resolution images with MATLAB bicubic kernel function to obtain low-resolution images. The original input size is 100x100 pixels, and we obtain low-resolution images by applying Integer down-sample factors from x2 to x8,which means total pixels reduce to 1/4 to 1/64. Figure ~\ref{fig_downsample} shows a few examples after down-sampling.

\subsection{Experiment Setup}
To compare the performance of ISR and FSR, we choose image super-resolution model Meta-SR\cite{hu2019meta} and RCAN\cite{zhang2018rcan} to compare with our method to evaluate the results with total accuracy. Note that for RCAN model, only results on x2,x3,x4,x8 are reported. Since Meta-SR model can generate high-resolution images at any size, we directly set the output size to 100x100. For RCAN, we use a different model for its corresponding down-sample factor, but for Meta-SR and our FSR-FER, we can use one model for all down-sample factors.
 \begin{figure}
 \centering
\includegraphics[width=10cm]{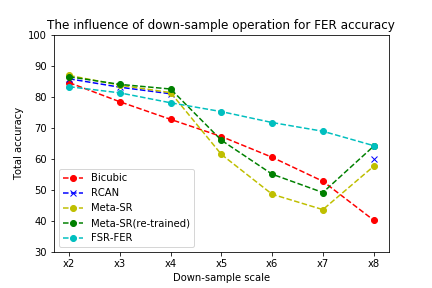}
\caption{Facial expression recognition accuracy comparisons at different down-sample scales.} \label{fig_res}
\end{figure}
\subsection{Facial Expression Recognition Results}
\begin{table}
\caption{Results of total accuracy on RAF-DB. Best results are {\bfseries highlighted}}\label{table1}
    \centering
    \setlength{\tabcolsep}{1mm}{
    \begin{tabular}{|c|c|c|c|c|c|c|c|}
    \hline
    \diagbox[innerwidth=3cm]{Method}{Total acc}{Scale}&x2&x3&x4&x5&x6&x7&x8\\ 
    \hline
    Bicubic & 84.518 & 78.357 & 72.686 & 67.177 & 60.495 & 52.771 & 40.264 \\
    \hline
    RCAN  & 85.756  & 83.051    & 80.932 &   -    &   -    &   -    & 59.909 \\
    \hline
    Meta-SR & {\bfseries 86.832} & 83.703 & 81.305 & 61.408 & 48.598 & 43.644 & 57.562 \\
    \hline
    Meta-SR (re-trained) & 86.408 & {\bfseries 83.963} & {\bfseries 82.464} & 65.971 & 55.052 & 49.153 & 64.003 \\
    \hline
    our FSR-FER & 83.214 & 81.226 & 78.031 & {\bfseries 75.196} & {\bfseries 71.675} & {\bfseries 68.872} & {\bfseries 64.244} \\
    \hline
    \end{tabular}}
\end{table}
Since our FSR-FER does not output high-resolution images, we compare directly on the total accuracy of facial expression recognition, and the results show in Figure~\ref{fig_res}. ISR methods has slightly better performance when resolution is relatively high. As the resolution decreases further, more image information is lost, and the performance of ISR methods has a significant decrease. Our proposed FSR-FER has better performance on down-sample factors larger than x4, which means feature-level super-resolution can better narrow the gap between features of high-ressolution images and generated features. Details are shown in Table ~\ref{table1}.

The approach of \cite{tan2018feature} utilizes feature-wise L2 loss as a stronger constraint to enable the generator to focus more on samples with larger loss. According to the idea of Focal loss\cite{lin2017focal}, the samples, which are easily misclassified, contributing more in the training process. So, we re-weight loss functions using the classification probability obtained by supervisor CovPoolFER model. The results in Table ~\ref{table2} shows that FSR-FER with classification-aware loss re-weighting has a slightly better performance. It is worth noting that it takes more than 400k iterations to achieve convergence without the introduction of loss re-weighting. But it takes about 150k iterations to achieve the same performance with proposed classification-aware loss re-weighting, which is about 3 times faster. 
\begin{table}
\caption{Result comparison of FSR-FER with and without loss re-weighting}\label{table2}
    \centering
    \setlength{\tabcolsep}{0.5mm}{
    \begin{tabular}{|c|c|c|c|c|c|c|c|c|}
    \hline
    \diagbox[innerwidth=2.5cm]{Method}{Total acc}{Scale}&x2&x3&x4&x5&x6&x7&x8&avg\\ 
    \hline
    w/o re-weighting& {\bfseries83.246} & 81.169 & {\bfseries78.196} &  74.739 &  71.675 & 68.774 &  64.211 & 74.573\\
    \hline
    w/ re-weighting & 83.214 & {\bfseries81.22} & 78.031 & {\bfseries 75.196} & 71.675 & {\bfseries 68.872} & {\bfseries 64.244} & {\bfseries74.637}\\
    \hline
    \end{tabular}}
\end{table}

\section{Discussion}
In this section, we discuss some details about performance difference between ISR methods and our FSR-FER. When the image resolution does not decrease much, ISR methods have better performance than FSR. A possible reason is that different from most ISR methods train each model for a single down-sample factor, FSR-FER is a one-for-all model, it is likely to focus on samples with more feature information loss.
 \begin{figure}
\includegraphics[width=\textwidth]{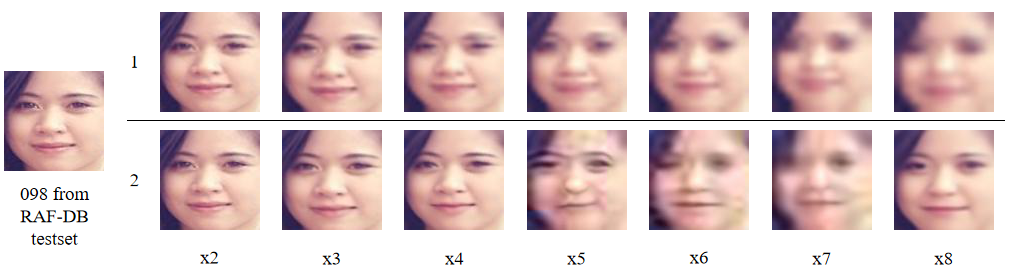}
\caption{Images in line 1 are the obtained by bicubic interpolation and line 2 are results of Meta-SR. } \label{fig_image}
\end{figure}
The resulting images of Meta-SR\cite{hu2019meta} shown in Figure ~\ref{fig_image}, it has excellent effects on down-sample factors of x2,x3 and x4, but not so good on x5, x6 and x7. We did not modify RCAN model to produce super-resolved images on x5, x6 and x7, but similar results were reported in \cite{tan2018feature} on VDSR\cite{kim2016vdsr} model and SRCNN model.

\section{Conclusion}
In this work, we proposed a novel feature super-resolution approach for facial expression recognition on images of various low resolutions. The main reason for the lower accuracy of low-resolution expression recognition is the loss of image feature information. Experiment results show our feature level super-resolution approach has a significant improvement on the accuracy of facial expression recognition on low-resolution images. Comparison with image super-resolution method suggests that our approach has better performance when image resolution is reduced at a relatively large ratio.

\section{Acknowledgement}
This work is supported by National Key Research and Development Program of China (2018YFB1004500), National Nature Science Foundation of China (61877048, 61472315), Innovative Research Group of the National Natural Science Foundation of China (61721002), Innovation Research Team of Ministry of Education (IRT 17R86), Project of China Knowledge Center for Engineering Science and Technology. Project of Chinese academy of engineering "The Online and Offline Mixed Educational Service System for 'The Belt and Road' Training in MOOC China".

%
%

%
%
%
\bibliographystyle{splncs04}
\bibliography{references}

\end{document}